\PassOptionsToPackage{numbers, compress}{natbib}

\documentclass{article}

\usepackage[preprint]{neurips_2025}

\usepackage[utf8]{inputenc} 
\usepackage[T1]{fontenc}    
\usepackage{hyperref}       
\usepackage{url}            
\usepackage{booktabs}       
\usepackage{amsfonts}       
\usepackage{nicefrac}       
\usepackage{microtype}      
\usepackage{xcolor}         
\usepackage{amsmath}
\usepackage{booktabs}
\usepackage{multirow}
\usepackage{graphicx}
\usepackage{colortbl}
\usepackage{xcolor}
\usepackage{makecell}
\usepackage[most]{tcolorbox}
\usepackage{wrapfig}
\usepackage{pifont}
\usepackage{float}
\usepackage{tabularx}
\usepackage{array}

\definecolor{ReasoningGreen}{HTML}{b0dfb9}
\definecolor{AssumptionRed}{HTML}{f39f9f}
\definecolor{UncertaintyGray}{HTML}{a2a4a5}

\title{CoT-Kinetics: A Theoretical Modeling Assessing \\LRM Reasoning Process}

\author {
Jinhe Bi\textsuperscript{\rm 1,}\textsuperscript{\rm 2} \quad
Danqi Yan\textsuperscript{\rm 1}\thanks{These authors contributed equally to this work. Email contact: \textit{bijinhe@outlook.com}.} \quad
Yifan Wang\textsuperscript{\rm 1}\footnotemark[1]  \quad
Wenke Huang\textsuperscript{\rm 3} \quad
Haokun Chen\textsuperscript{\rm 1} \quad
Guancheng Wan\textsuperscript{\rm 3} \quad \\
\textbf{Mang Ye\textsuperscript{\rm 3} \quad
Xun Xiao\textsuperscript{\rm 2}}\thanks{Corresponding authors: \textit{cognitive.yunpu@gmail.com}, \textit{drxiaoxun@gmail.com}.} \quad  
\textbf{Hinrich Schütze\textsuperscript{\rm 1,}\textsuperscript{\rm 4} }\quad 
\textbf{Volker Tresp\textsuperscript{\rm 1,}\textsuperscript{\rm 4} \quad 
Yunpu Ma\textsuperscript{\rm 1,}\textsuperscript{\rm 4}\footnotemark[2]} \\
\textsuperscript{\rm 1} Ludwig Maximilian University of Munich \quad \textsuperscript{\rm 2} Munich Research Center, Huawei Technologies \\
\textsuperscript{\rm 3} School of Computer Science, Wuhan University \quad 
\textsuperscript{\rm 4} Munich Center for Machine Learning
}

\begin{document}

\maketitle

\begin{abstract}
Large Reasoning Models (LRMs) significantly improve the reasoning ability of Large Language Models (LLMs) by learning to reason, exhibiting the promising performance in solving complex tasks.
LRMs solve a task that require complex reasoning by explicitly generating a chain‑of‑thought (CoT) reasoning trajectory before concluding an answer. 
Nevertheless, judging the quality of such an output answer is not easy because only considering the correctness of the answer is not enough and the soundness of the reasoning trajectory part matters as well. Logically, if the soundness of the reasoning part is poor, even if the answer is correct, the confidence of the derived answer should be low.
Existing methods did consider a joint assessment with taking into account the reasoning part, however, their precision is unsatisfactory as the causal relationship of the reasoning to the concluded answer still cannot properly reflected.
In this paper, inspired by classical mechanics, we present a novel approach towards establishing a \textit{CoT-Kinetics energy} equation for the reasoning process. Specifically, our CoT-Kinetics energy equation formulates the token state transformation process regulated by LRM internal transformer layers, as like a particle kinetics dynamics  governed in a mechanical field. 
Our CoT-Kinetics energy assigns a scalar score to  evaluate particularly the soundness of the reasoning phase, telling how confident the derived answer could be based on the evaluated reasoning. As such, the LRM's overall output quality can be measured with finer granularity, rather than a coarse judgment (e.g., correct or incorrect) anymore. 
We comprehensively evaluated the fidelity of the CoT-Kinetics energy modeling. Results justify that our CoT-Kinetics energy score indeed logically reflects the causal relationship of the reasoning part and the derived final answer, outperforming  existing baselines in terms of assessment metrics of AUROC, AUPR and FPR@95, across seven open-source LRMs and six widely recognized benchmarks.
Beyond that, our work shows a potential to assist LRMs to build a feedback loop to improve its reasoning process by judging the quality of its output answer based on our CoT-Kinetic energy score.
\end{abstract}
\section{Introduction}

\begin{figure}[t]
    \centering
    \includegraphics[
        width=1\linewidth
    ]{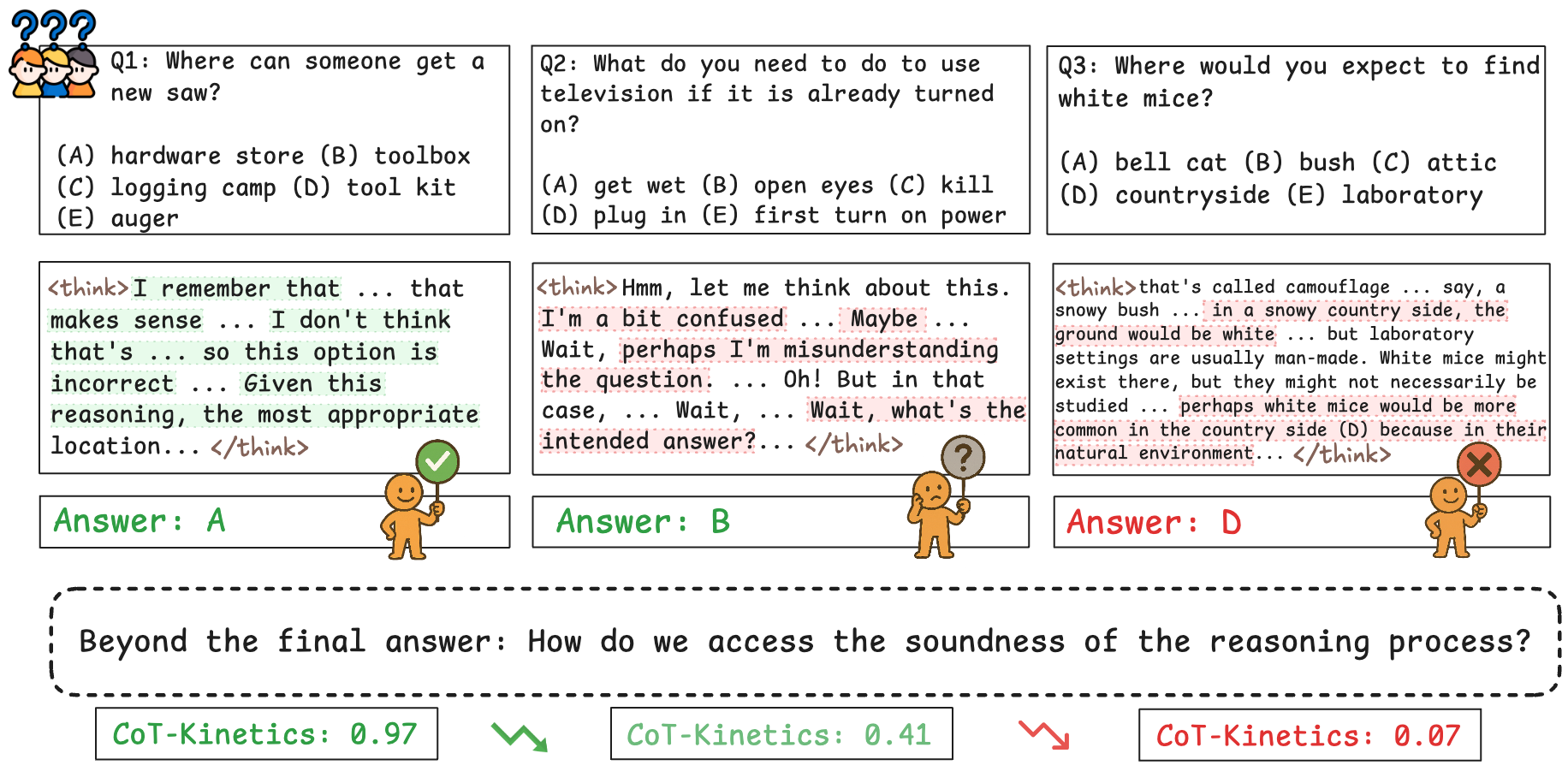}
    \caption{
Illustration of the necessity to separately assess the reasoning trajectory part with fine-grained granularity. 
Given a query from a specific domain, an LRM generates both a final answer and a CoT reasoning trajectory. 
However, the final answer may still be correct even if the reasoning path contains \textcolor{AssumptionRed}{flawed assumptions or unstable logic}.
As shown in Example~2, evaluating only the final output overlooks deficiencies in the reasoning process.
A higher score (Example~1) reflects a \textcolor{ReasoningGreen}{more sound reasoning trajectory} and a greater likelihood of correctness, whereas a lower score (Example~3) indicates \textcolor{AssumptionRed}{flawed assumptions or unstable logic} and a higher chance of error. It motivates that only considering the correctness of the final answer is inappropriate while an accurate assessment technique such as the proposed CoT-Kinetics energy is required.
}
\label{fig:Fig1}
\end{figure}

Large Reasoning Models (LRMs) \cite{deepseekai2025deepseekr1incentivizingreasoningcapability,openai2024openaio1card,qwq32b} show significant improvements on solving complex reasoning tasks \cite{chen2023theoremqatheoremdrivenquestionanswering,cobbe2021gsm8k,hendrycks2021measuringmassivemultitasklanguage,shi2022languagemodelsmultilingualchainofthought,wang2024mmluprorobustchallengingmultitask,talmor-etal-2019-commonsenseqa} by generating an output answer with a structured and deliberate reasoning trajectory. However, evaluating the quality of an LRM's final output answer is inherently complex~\citep{wu2024oceanofflinechainofthoughtevaluation,taubenfeld2025confidenceimprovesselfconsistencyllms}. It has to judge the causal relationship of the reasoning trajectory to the concluded answer because a correct answer does not necessarily mean that its reasoning trajectory is coherent and sound~\citep{nguyen2024directevaluationchainofthoughtmultihop, wang2024chain, turpin2023languagemodelsdontsay}. This situation is illustrated in the Example 2 (middle case) in Figure~\ref{fig:Fig1}, where the reasoning trajectory is poor but the derived final answer is still correct. Therefore, accurately assessing the reasoning soundness is crucial for judging the quality of an output answer generated by an LRM.

Existing methods in literature did consider how to assess the causal relationship of the reasoning part and the derived final answer. Those methods can be categorized into  two types.
One school relies on external assessment techniques, such as supervision from human annotations, linear probing techniques, or auxiliary judge models~\citep{wang2025thoughtprobe, li2023makinglargelanguagemodels, he2025largelanguagemodelsdetect}. 
Another line of work models output reliability by quantifying uncertainty (e.g., via softmax confidence, temperature scaling) or analyzing semantic consistency (e.g., hidden state entropy) of internal model states without external supervision~\citep{wang2025latentspacechainofembeddingenables, shih2023longhorizontemperaturescaling, Huang_2025, si2023promptinggpt3reliable}. A more detailed summary of each type is provided in Appendix~\ref{rel}.

On the one hand, the problem of the first type of methods is that although these approaches are effective under supervised evaluation environments, they suffer from limited generality and become impractical in real-world scenarios, because the external information is not always available; in addition, the query distributions may also deviate significantly from training environments. On the other hand, the key shortcoming of the second type of methods is that though their modeling approach is promising, the theoretical modeling technique is inadequate. Also, the second type of methods evaluate the entire output without splitting the reasoning trajectory from the overall output answer. This makes the correctness of the final answer inevitably dominating the level of soundness of the reasoning phase, thus the  causal relationship is not properly reflected.

In this paper, inspired by classical mechanics, we propose a novel theoretical modeling, \textit{CoT-Kinetics} energy, along the line of the second type of methods. Different to the existing internal modeling methods~\citep{wang2025latentspacechainofembeddingenables, shih2023longhorizontemperaturescaling, Huang_2025, si2023promptinggpt3reliable,plaut2025probabilitieschatllmsmiscalibrated,hendrycks2022scalingoutofdistributiondetectionrealworld,kuhn2023semanticuncertaintylinguisticinvariances,duan2024shiftingattentionrelevancepredictive}, We analogize the state transition of reasoning tokens as like a particle kinetics dynamics governed in a mechanical field, under a ``force field'' regulated by LRM's transformer layers.
Our modeling aims to construct an energy equation that can represent the soundness level of the reasoning process. The proposed CoT-Kinetics energy formulates the intrinsic semantic dynamics of the model's internal reasoning process as a form of kinetic energy, taking into account the uncertainty induced by the query task. At the end, the CoT-Kinetics energy assigns a scalar score to evaluate the soundess of a given reasoning trajectory. The details of the CoT-Kinetics energy modeling will be elaborated in Section~\ref{Methodology}.

First of all, our modeling approach solely relies on internal state information during the reasoning process of the LRM. This overcomes the shortcomings of the first type of methods previously discussed, where supervised information is needed. Second, our solution particularly focuses the reasoning part and explicitly assesses its soundness alone; besides, our solution is built with a principled modeling technique where a CoT-Kinetics energy equation is established for the reasoning process. This fundamentally evolves the capability of the second type of methods. Later, we will see that the our CoT-Kinetics energy modeling can accurately interpret the causal relationship of the reasoning trajectory to the derived final answer. Therefore, the CoT-Kinetics energy score is an precise indicator to the quality of the overall output answer for LRMs.

Comprehensive evaluations have been done to judge the performance of the proposed solution. Specifically, three metrics (AUROC, AUPR, and FPR@95) are selected to measure how satisfactory the CoT-Kinetics energy score can reflect the soundness of the reasoning trajectory and do match to the correctness of the final answer. The experiments were done across seven open-source LRMs \cite{deepseekai2025deepseekr1incentivizingreasoningcapability,openai2024openaio1card,qwq32b} and six diverse benchmarks \cite{chen2023theoremqatheoremdrivenquestionanswering,cobbe2021gsm8k,hendrycks2021measuringmassivemultitasklanguage,shi2022languagemodelsmultilingualchainofthought,wang2024mmluprorobustchallengingmultitask,talmor-etal-2019-commonsenseqa} covering mathematics, commonsense reasoning, theorem proving, general knowledge retrieval, and multilingual understanding. Empirical results demonstrate that CoT-Kinetics consistently achieves superior performance compared to existing baselines under varied scenarios.

Our contributions can be summarized as follows:
\begin{itemize}
    \item We introduce a theoretical modeling dedicated for the LRM reasoning phase by framing the internal semantic evolution as a discrete-time dynamics system, enabling a structured analysis of layer-wise reasoning process within LRM's transformer.
    \item Based on the modeling, inspired with classical mechanics, we establish a \textit{CoT-Kinetics} energy equation where the intrinsic semantic dynamics of state changes of reasoning tokens is characterized as a form of kinetic energy, with taking into account the uncertainty induced by the query task. This energy equation gives a scalar score to the soundness of the reasoning part of an LRM and can be used as an indicator to measure the quality of the overall output answer.
    \item Comprehensive  evaluations are conducted across seven open-source LRMs and six challenging benchmarks. Results show that the proposed CoT-Kinetic energy modeling is indeed a better modeling approach comparing with the state-of-the-art methods.
\end{itemize}

\vspace*{-0.2em}
\section{CoT-Kinetics Energy: Modeling LRM CoT Reasoning Process}
\vspace*{-0.2em}
\label{Methodology}
\subsection{Formulation of LRM CoT Reasoning}

LRMs unfold reasoning as a structured and deliberative process, where the model generates not only a conclusive answer but also a reasoning trajectory together as the outcome of inference, as the examples shown in Figure~\ref{fig:Fig1}.
Upon receiving a query task, an LRM will extract and encode the information contained in the question into a unified representation to facilitate the subsequent answering with thinking processes. Following the query embedding processing, the model starts its reasoning phase, where it generates intermediate reasoning tokens between designated special tokens such as \texttt{<think>} and \texttt{</think>}, which will be eventually translated into textual information as the reasoning trajectory.

For this specific reasoning phase, we assume the total sequence length of generated CoT reasoning tokens is \(K\), wherein the \(k\)-th generated reasoning token is denoted as \(q_k\). 
We further formalize the layer-wise inference process in the transformer for generating \(q_k\) as follows. At each transformer layer, the hidden state of each token is updated based on the model parameters and the accumulated processing results from the previous layers.
Formally, let \(q_k^i \in \mathbb{R}^d\) denote the hidden state of the \(k\)-th token modified after the \(i\)-th layer computation, where \(d\) is the hidden dimension. The state transition from \(q_k^{i-1}\) to \(q_k^{i}\) can be represented as a transformation function below:
\begin{equation}
q_k^{i} = F_i(q_k^{i-1}, W^{i}),~\forall i=1,2,\cdots,L~,
\label{eq:layer_transformation}
\end{equation}
where \(W^i\) represents the synthesized effect of the model parameters and attention mechanisms applied to the \(q_k^{i-1}\); and \(L\) is the total number of transformer layers of an LRM.

Once the final reasoning token state \(q_K^L\) is generated, the reasoning phase is considered complete, and the model proceeds to the next phase of generating the final answer, given the resulting reasoning trajectory.
In this paper, as mentioned, we focus on the reasoning phase and will next model the reasoning process with an energy equation.

\subsection{CoT-Kinetics Energy Equation}

\subsubsection{Intermediate State Information}

Modeling the reasoning process first relies on the intermediate states of the generated reasoning tokens collected during the reasoning phase. These reasoning tokens are organized into a \((K, L, d)\)-dimensional tensor.
In addition to the hidden states of reasoning tokens, we also in parallel collect information about the model’s internal uncertainty at each inference step by measuring the entropy of the output logits corresponding to each generated token. 
This entropy signal reflects the model's epistemic uncertainty during the reasoning process and indirectly characterizes the inherent difficulty posed by the query~\cite{Huang_2025}. This metric is widely used for such a purpose in previous works~\cite{Huang_2025,wang2025latentspacechainofembeddingenables}.
These intermediate information lays the foundation for establishing the CoT-Kinetics energy equation.

\subsubsection{Formulation of CoT-Kinetics Energy}\label{sssec:formulCoTK}
The state transition of a reasoning token happens layer-by-layer in the transformer of an LRM, regulated by the model parameters at each layer as in Eq.~(\ref{eq:layer_transformation}). Such a state transition can be imagined as a particle traveling through a force field in classical mechanics, if we view the reasoning token \(q_k^i\) being a particle and consider the LRM layer parameters \(W^i\) being the force field, applied to change the reasoning token state in a discrete time manner. This analogy inspires us to establish a kinetic energy equation for the LRM CoT reasoning process.

We first introduce the pre-processing stage before presenting the CoT-Kinetic energy equation. We first apply mean pooling over the collected hidden states that are modified by the same transformer layer \(i\) across all reasoning tokens. The mean pooling operation is done by averaging all \(K\) reasoning tokens  generated by the same layer \(i\), wherein:
\begin{equation}
\bar{q}^i = \frac{1}{K} \sum_{k=1}^{K} q_k^i~,
\label{eq:reasoning_mean_pooling}
\end{equation}
where \(\bar{q}^i\) serves as a compact semantic summary of the reasoning trajectory at the \( i \)-th layer, which aggregates the representational states of all intermediate reasoning steps excluding the final answer tokens. 

The main purpose of doing the mean pooling above is explained as follows. First of all, since these tokens are generated within the dedicated reasoning phase (i.e., between \texttt{<think>} and \texttt{</think>}), their averaged representation preserves the core semantics without information loss. 
Moreover, mean pooling operation reduces token-level variance and highlights the global progression of reasoning across layers, which is critical for modeling the layer-wise dynamics. Note that this simplification strategy has been widely adopted in prior work for semantic representation modeling~\cite{gao2021simcse, reimers2019sentencebertsentenceembeddingsusing, bi2025prismselfpruningintrinsicselection}.
Based on the averaged states \(\bar{q}^i\) in Eq.~(\ref{eq:reasoning_mean_pooling}), next we model the semantic evolution of the aggregated reasoning state across transformer layers as a discrete particle dynamics process, under the influence of the LRM layer parameters \( W^i \) as defined in Eq.~\ref{eq:layer_transformation} (conceptualized here as internal forces), 

In classical mechanics, kinetic energy is determined by momentum, while curvature characterizes the geometry of the particle's trajectory.~\cite{arreagagarcia2013equationsmotionrelativisticcharged, hsiao2024geometricallyconstrainedparticledynamics}. 
Analogously, to define the total semantic kinetic energy, we introduce two quantities as follows: 
\begin{itemize}
    \item \(\tau^i\): \textit{Semantic Momentum Energy} 
 quantifying the magnitude of semantic displacement between two consecutive layers, wherein: 
    \begin{equation}
\tau^i = \|\bar{q}^i - \bar{q}^{i-1}\|~.
\label{eq:mom_tau}
\end{equation}
    \item \(\kappa^i\): \textit{Semantic Curvature Energy} 
 capturing the second-order variation of semantic transitions across layers, wherein:
    \begin{equation}
\kappa^i = \|(\bar{q}^i - \bar{q}^{i-1}) - (\bar{q}^{i-1} - \bar{q}^{i-2})\|~.
\label{eq:curv_kappa}
\end{equation}

\end{itemize}

The interpretation of the two quantities in the context of reasoning process is explained as follows. \(\tau^i\) measures the extent of semantic progression, encouraging the reasoning trajectory to continuously advance across layers, while \(\kappa^i\) captures the degree of semantic correction, reflecting the model's ability to dynamically adjust its reasoning steps.
For doing a better reasoning, both strong progression and active correction are desirable: large \(\tau^i\) indicates that the model is making meaningful semantic exploration rather than stagnating, and large \(\kappa^i\) suggests that the model is flexibly refining its trajectory rather than following a rigid or erroneous path. Thus, by considering these two aspects, our CoT-Kinetics energy can better represent the internal reasoning process.

With the above definitions, finally, we establish the CoT-Kinetics energy \( \mathcal{E}_\text{CoT} \) equation as:
\begin{equation}
\mathcal{E}_\text{CoT} = \sum_{i=0}^{L-1} \left( \tau^i + \kappa^i \right)- \gamma \cdot\mathcal{H}(p)  
~,
\label{eq:cot_kinetics_energy}
\end{equation}
where \(\gamma\) is a hyperparameter used as a scaling factor to align the magnitude of the entropy term \(\mathcal{H}(p)\) with the kinetic energy components, and \(\tau^i\) and \(\kappa^i\) were already specified in Eq.~(\ref{eq:mom_tau}) and Eq.~(\ref{eq:curv_kappa}). Note that both \(\tau^i\) and \(\kappa^i\) are normalized by the total displacement between the initial and final reasoning representations \((\bar{q}^0, \bar{q}^{L-1})\) to ensure scale-invariant dynamics across varying input complexities. 

The CoT-Kinetics energy \(\mathcal{E}_\text{CoT}\) defined in Eq.~(\ref{eq:cot_kinetics_energy}) is a scalar value that assigns a score to a given reasoning process. This can quantitatively assess the soundness of the generated reasoning tokens. Again, the whole CoT-Kinetics energy \(\mathcal{E}_\text{CoT}\) can be built solely with the internal intermediate token states, without any external information needed as in prior art \cite{shih2023longhorizontemperaturescaling, Huang_2025, si2023promptinggpt3reliable,wang2025latentspacechainofembeddingenables}.

\subsection{Remarks}

For a better understanding, the proposed CoT-Kinetics energy can be interpreted as follows. Intuitively, a higher \( \mathcal{E}_\text{CoT}\) score can be interpreted as a reasoning trajectory that undergoes large and rapid semantic exploration thus successfully overcoming semantic potential barriers (i.e., achieving lower entropy) imposed by the query task. Such a reasoning trajectory is hypothesized to demonstrate better reasoning soundness and are thus more likely to conclude with a correct answer.

Our modeling method offers several key features:
(i) it requires no external supervision, operating in a fully training-free manner and no need to possess ground-truth labels;
(ii) it is grounded in internal representational dynamics, making it inherently interpretable and adaptable across different model architectures; and (iii) it can be seamlessly integrated into existing LLM inference workflows without requiring any architectural modifications.

\vspace*{-0.2em}
\section{Experiments}
\vspace*{-0.2em}
We begin by detailing the experimental setup, and then present the comparison results of the proposed approach using the energy score \(\mathcal{E}_\text{CoT}\) against selected baselines. 
Subsequently, we provide a comprehensive analysis of the effectiveness of our method in evaluating reasoning quality across multiple dimensions, including domain specificity versus generality, model scale variation, and multilingual scenarios. 
Finally, we conduct an ablation study to investigate the effects of different reasoning token aggregation strategies and semantic energy component designs on the overall evaluation performance.

\subsection{Experiment Setup}

\begin{figure}[t]
    \centering
    \begin{minipage}[t]{0.61\linewidth}
        \centering
        \includegraphics[width=\linewidth, trim={0 540 840 0}, clip]{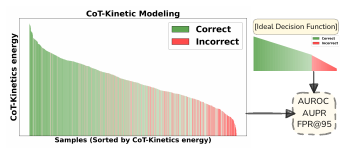}
        \label{fig:ranking}
    \end{minipage}
    \hfill
    \begin{minipage}[t]{0.36\linewidth}
        \centering
        \includegraphics[width=\linewidth]{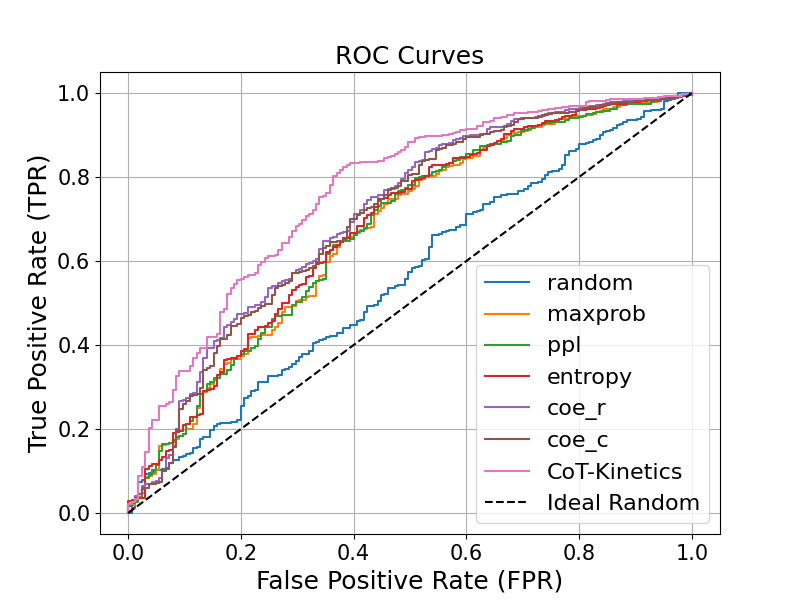}
        \label{fig:roc_curves}
    \end{minipage}
\caption{
Left: Samples ranked by CoT-Kinetics energy \( \mathcal{E}_\text{CoT} \), where higher scores align with correct outputs, reflecting the soundness of reasoning and logical causal relationship in terms of AUROC, AUPR, and FPR@95 metric.
Right: ROC curves comparing CoT-Kinetics energy \( \mathcal{E}_\text{CoT} \) with baselines, where the dashed line denotes the expected performance of a random scoring function as a lower bound.
}
\label{fig:cot-kinetics-eval}
\end{figure}

\textbf{Assessment Criteria:} 
If our proposed CoT-Kinetic energy modeling is satisfactory, intuitively, the energy score \(\mathcal{E}_\text{CoT}\) should quantify the level of soundness of the reason trajectory and objectively reflect the causal relationship of the reasoning to the derived answer. This means that if the reasoning trajectory is considered highly confident (i.e., yielding a high CoT-Kinetics energy score), then the derived final answer shall be correct with high probability. Oppositely, if the energy score \(\mathcal{E}_\text{CoT}\) often contradicts to the fact of the final answer, then the CoT-Kinetic energy modeling is considered unsatisfactory.
Measuring the severity of dissatisfaction is equivalent to a ranking  problem: as illustrated in  Figure~\ref{fig:cot-kinetics-eval} (Left part), the lower the probability that an incorrect sample (red) appears in the high-score region, the better our energy modeling satisfies the desired assessment criterion.
Technically, such a severity can be quantitatively measured. In this paper, we use three existing metrics to interpret how satisfactory our proposed CoT-Kinetics energy modeling fulfills the criteria. The definition of the three metrics are introduced in the next subsection.

\textbf{Assessment Metrics:} 
The three metrics used for evaluation are AUROC, AUPR and FPR@95. As mentioned, the three metrics build a relationship between the soundness level of a reasoning trajectory to the correctness of the answer but from different perspectives with different means. Each metric captures a different assessment aspect:
\begin{itemize}
    \item \textit{AUROC}: a metric focusing on the global correctness ranking capability. Higher AUROC means that the proposed CoT-Kinetics energy modeling method consistently assigns higher energy scores to more reasonable reasoning trajectories.    
    \item \textit{AUPR}: a metric emphasizing the precision at identifying correct answers. It assesses the capability of whether or not the CoT-Kinetics energy modeling method can prioritize truly sound reasoning trajectory under correctness imbalance. A higher AUPR indicates a stronger ability to accurately distinguish reliable reasoning even when correct cases are rare.    
    \item \textit{FPR@95}: a metric evaluating the robustness at high recall. A lower FPR@95 score indicates that the CoT-Kinetics energy modeling method reliably identifies correct reasoning trajectories with minimal false positives, even under stringent correctness requirements.
\end{itemize}
These three metrics are widely adopted in the machine learning community as standard tools for evaluating ranking-based binary classification performance, particularly under imbalanced or high-recall conditions. Individual definitions of the three metric can be found in Appendix~\ref{appendix:evaluation_metrics}. The metrics offer a multi-faceted, rigorous evaluation of how logical the CoT-Kinetics energy score can reflect the soundness of the underlying reasoning given the fact of the correctness of the final answer.

\textbf{Baselines:}
We choose six baselines for comparison: \texttt{Random}, \texttt{Maximum Softmax Probability}~\cite{wang2025latentspacechainofembeddingenables}, \texttt{Perplexity}~\cite{si2023promptinggpt3reliable}, \texttt{Entropy}~\cite{Huang_2025}, and two recent representation-based scoring methods, \texttt{CoE-R} and \texttt{CoE-C}~\cite{wang2025latentspacechainofembeddingenables}. For each method, we compute and report all three evaluation metrics described above.
In particular, the \texttt{Random} baseline serves as a lower bound, representing a random ranking method with no evaluation capability.

\textbf{Models and Benchmarks:}
We evaluate the proposed CoT-Kinetics energy modeling/scoring on seven state-of-the-art LRMs with different model scales, including \texttt{DeepSeek-R1-Distill-Qwen-1.5B/7B/14B/32B}, \texttt{DeepSeek-R1-Distill-Llama-8B/70B} \cite{deepseekai2025deepseekr1incentivizingreasoningcapability}, and \texttt{QwQ-32B} \cite{qwq32b}, each exhibiting varying levels of reasoning proficiency. Detailed model architectures and training configurations are provided in Appendix~\ref{appendix:model-configurations}.

To demonstrate the generality of the proposed CoT-Kinetics energy modeling, we conduct experiments on six widely adopted reasoning benchmarks. MMLU and MMLU-Pro provide multi-domain coverage across a broad range of subjects requiring complex reasoning. GSM8K, TheoremQA, and CommonsenseQA target distinct reasoning capabilities, including arithmetic problem solving, formal logical inference, and commonsense understanding. In addition, we evaluate on the multilingual benchmark MGSM, using English, German, French, and Chinese subsets to assess cross-lingual applicability. Further details of each benchmark dataset, including task formulations, prompts, and example questions, are provided in Appendix~\ref{appendix:dataset_details}.

\subsection{Main Results}
\begin{table*}[t]
\centering
\normalsize
\setlength{\tabcolsep}{2pt}  
\renewcommand{\arraystretch}{1.1}  
\caption{Main results on GSM8K, CommonsenseQA, and TheoremQA. Arrows indicate direction of improvement.}
\resizebox{\textwidth}{!}{%
\begin{tabular}{ll*{9}{c}}
\toprule
\multirow{2}{*}{Model} & \multirow{2}{*}{Method} 
& \multicolumn{3}{c}{\textbf{GSM8K}} 
& \multicolumn{3}{c}{\textbf{CommonsenseQA}} 
& \multicolumn{3}{c}{\textbf{TheoremQA}} \\
\cmidrule(lr){3-5} \cmidrule(lr){6-8} \cmidrule(lr){9-11}
& & AUROC~↑ & AUPR~↑ & FPR@95~↓ 
  & AUROC~↑ & AUPR~↑ & FPR@95~↓ 
  & AUROC~↑ & AUPR~↑ & FPR@95~↓ \\
\midrule

\multirow{7}{*}{\makecell[l]{R1-Qwen-1.5B}} 
& Random         & 50.23 & 75.39 & 94.29 & 50.03 & 46.81 & 93.14 & 49.75 & 40.35 & 92.81 \\
& MaxProb   & 70.34 &87.97& 76.95 &64.06&61.16&81.42&53.65&43.91&92.19\\
& PPL       &70.88&88.24&77.29&63.86&61.07&83.19&54.05&44.24&91.25\\
& Entropy   &71.86&88.54&75.93&64.26&61.32&82.30  &54.33&44.52&92.19\\
& CoE-R     &74.17&89.99&75.93&60.03&56.37&85.18&53.33&44.18&93.44\\
& CoE-C     &73.32&89.37&75.25&58.90&55.44&85.18&52.42&43.34&90.31\\
\rowcolor{blue!6}
& CoT-Kinetics & \textbf{77.28} & \textbf{91.42} & \textbf{72.20} & \textbf{64.26} & \textbf{61.34} & \textbf{81.19} & \textbf{54.37}  & \textbf{44.82} & \textbf{90.00}\\
\midrule

\multirow{7}{*}{\makecell[l]{R1-Qwen-7B}} 
& Random         & 53.56 & 85.88 & 93.20 & 48.51 & 71.16 & 98.02 & 46.39 & 48.46 & 98.12 \\
& MaxProb   &67.05&92.26&82.42&76.64&86.89&67.46&48.95&49.88&89.85\\
& PPL       &67.27&92.32&82.42&76.41&88.33&67.46&49.38&50.16&91.35\\
& Entropy   &68.04&92.54&79.39&74.72&88.03&68.25&49.58&50.31&90.98\\
& CoE-R     &70.95&92.98&73.94&71.91&85.58&75.79&50.99&53.31&90.60\\
& CoE-C     &70.12&92.74&74.55&70.42&84.81&76.98&50.56&52.94&90.23\\

\rowcolor{blue!6}
& CoT-Kinetics & \textbf{76.32} & \textbf{94.73} & \textbf{67.27}& \textbf{77.41}  & \textbf{88.51} & \textbf{65.87} & \underline{50.65} & 52.49 & \textbf{90.23} \\
\midrule

\multirow{7}{*}{\makecell[l]{R1-LLaMA-8B}} 
& Random         & 50.22 & 81.21 & 95.33 & 51.75 & 77.22 & 96.14  & 46.29 & 41.67 & 92.91 \\
& MaxProb   & 58.85&85.46&94.49&74.71&89.29&67.38  & 59.39 &53.12&84.65 \\
& PPL       &58.94&85.51&94.49&74.55&89.23&67.81  &60.01&53.91&85.04\\
& Entropy   &60.25&85.89&94.88&74.68&89.28&65.24  &60.27&54.06&84.65\\
& CoE-R     &63.01&86.57&86.22&52.00&77.92&93.56  &50.56&46.02&96.46\\
& CoE-C     &63.86&86.50&86.61&73.90&88.92&69.96  &58.01&54.10&93.70\\

\rowcolor{blue!6}
& CoT-Kinetics & \textbf{68.12} & \textbf{89.37} & \textbf{80.31} & \textbf{74.71} & \textbf{89.35} & \textbf{65.24} & \textbf{60.57} & \textbf{56.06} & \textbf{83.86} \\
\midrule

\multirow{7}{*}{\makecell[l]{R1-Qwen-14B}} 
& Random         & 49.46  & 93.23 & 95.24 & 49.13 & 84.30 & 95.03 & 50.93 & 56.79 & 94.93 \\
& MaxProb   &53.98&93.06&85.71&78.90&94.87  &65.22&51.94&57.71&89.40\\
& PPL       &54.51&93.23&84.52&79.24&94.99&64.60&52.39&58.07&89.40\\
& Entropy   &55.13&93.26&85.71&79.24&94.97&65.84&52.62&58.24&90.32\\
& CoE-R     &59.62&93.84&77.38&71.78&91.82&73.29&52.28&58.32&93.55\\
& CoE-C     &58.63& 93.66 & 77.38 &68.40&90.78&80.75&50.99&57.39&92.63\\

\rowcolor{blue!6}
& CoT-Kinetics & \textbf{61.64} & \textbf{94.04} & \textbf{76.19} & \textbf{79.27} & \textbf{95.00} & \textbf{61.49} & \textbf{52.65}& \textbf{58.77}  & \textbf{88.02} \\

\bottomrule
\end{tabular}
}
\label{tab:main_results_3bench}
\end{table*}
\begin{table*}[t]
\centering
\normalsize
\setlength{\tabcolsep}{2pt}
\renewcommand{\arraystretch}{1.1}
\caption{
Performance comparison on MMLU and MMLU-Pro, two multi-task benchmarks requiring both world knowledge and complex reasoning. Arrows indicate direction of improvement}
\resizebox{\textwidth}{!}{%
\begin{tabular}{ll|*{12}{c}}
\toprule
\multirow{2}{*}{Dataset} & \multirow{2}{*}{Method} 
& \multicolumn{3}{c}{\textbf{R1-Qwen-1.5B}} 
& \multicolumn{3}{c}{\textbf{R1-Qwen-7B}} 
& \multicolumn{3}{c}{\textbf{R1-LLaMA-8B}} 
& \multicolumn{3}{c}{\textbf{R1-Qwen-14B}} \\
\cmidrule(lr){3-5} \cmidrule(lr){6-8} \cmidrule(lr){9-11} \cmidrule(lr){12-14}
& & AUROC~↑ & AUPR~↑ & FPR@95~↓ 
  & AUROC~↑ & AUPR~↑ & FPR@95~↓ 
  & AUROC~↑ & AUPR~↑ & FPR@95~↓ 
  & AUROC~↑ & AUPR~↑ & FPR@95~↓ \\
\midrule

\multirow{7}{*}{MMLU-Pro}
& Random         & 49.51 & 25.83 & 95.72 & 50.71 & 30.83 & 94.96 & 50.95 & 31.25 & 94.30 & 50.22 & 40.12 & 94.31 \\
& MaxProb        & 61.44 & 33.40 & 91.05 & 60.11 & 35.55 & 91.87 & 59.29 & 36.68 & 90.54 & 65.54 & 48.33 & 85.52 \\
& PPL            & 61.65 & 33.38 & 90.89 & 60.31 & 35.86 & 90.96 & 59.42 & 36.86 & 90.44 & 65.34 & 47.98 & 86.23 \\
& Entropy        & 62.28 & 33.93 & 90.57 & 59.89 & 34.89 & 90.10 & 59.82 & 37.12 & 89.93 & 66.12 & 48.31 & 85.45 \\
& CoE-R          & 57.78 & 30.76 & 91.25 & 52.63 & 31.34 & 92.81 & 51.77 & 33.66 & 94.12 & 61.45 & 45.12 & 86.77 \\
& CoE-C          & 54.50 & 28.45 & 92.95 & 55.41 & 32.89 & 92.13 & 57.89 & 36.20 & 89.60 & 61.52 & 46.53 & 87.10 \\
& CoT-Kinetics 
                & \cellcolor{blue!6}\textbf{62.46} & \cellcolor{blue!6} 33.39 & \cellcolor{blue!6}\textbf{90.41} 
                & \cellcolor{blue!6}\textbf{61.89} & \cellcolor{blue!6}\textbf{36.13} & \cellcolor{blue!6}\textbf{89.45} 
                & \cellcolor{blue!6}\textbf{60.47} & \cellcolor{blue!6}\textbf{37.90} & \cellcolor{blue!6}\textbf{89.24} 
                & \cellcolor{blue!6}\underline{65.89} & \cellcolor{blue!6}\textbf{48.57} & \cellcolor{blue!6}\textbf{85.12} \\
\midrule

\multirow{7}{*}{MMLU}
& Random         & 50.22 & 57.23 & 95.80 & 48.27 & 70.83 & 95.42 & 50.29 & 78.96 & 95.75 & 50.65 & 85.72 & 94.69 \\
& MaxProb        & 64.61 & 75.37 & 91.81 & 72.11 & 87.04 & 84.77 & 71.29 & 89.55 & 82.54 & 78.20 & 93.77 & 71.69 \\
& PPL            & 65.19 & 75.50 & 90.96 & 72.33 & 87.22 & 84.36 & 71.32 & 89.68 & 83.25 & 78.26 & 94.03 & 72.28 \\
& Entropy        & 66.40 & 76.38 & 89.95 & 73.23 & 87.65 & 83.62 & 71.87 & 89.87 & 83.16 & 78.17 & 93.73 & 72.42 \\
& CoE-R          & 58.79 & 65.18 & 93.20 & 64.94 & 80.61 & 85.79 & 49.49 & 80.25 & 95.31 & 71.59 & 92.94 & 82.95 \\
& CoE-C          & 56.34 & 62.21 & 93.97 & 62.74 & 79.32 & 88.27 & 66.34 & 86.60 & 85.83 & 69.34 & 92.39 & 87.02 \\
& CoT-Kinetics 
                & \cellcolor{blue!6}\textbf{68.14} & \cellcolor{blue!6}\textbf{76.47} & \cellcolor{blue!6}\underline{90.36}
                & \cellcolor{blue!6}\textbf{74.26} & \cellcolor{blue!6}\textbf{88.18} & \cellcolor{blue!6}84.39 
                & \cellcolor{blue!6}\textbf{73.49}  & \cellcolor{blue!6}\textbf{90.71}  & \cellcolor{blue!6}\underline{82.59}  
                & \cellcolor{blue!6}\textbf{79.13}  & \cellcolor{blue!6}\textbf{94.84}  & \cellcolor{blue!6} \underline{72.12}  \\

\bottomrule
\end{tabular}
}
\label{tab:mmlu}

\end{table*}
We present the comprehensive evaluation on our CoT-Kinetics energy modeling approach across multiple experimental settings. 
First, we investigate the generality of CoT-Kinetics energy modeling across different reasoning domains and task difficulties, including mathematical reasoning, commonsense reasoning, and multi-task knowledge reasoning in Table \ref{tab:main_results_3bench} and Table \ref{tab:mmlu}. 
Table~\ref{tab:gsm8k_large_models} further reports results on larger-scale models, including two 32B models and one 70B model, illustrating 
the scalability of CoT-Kinetics to more powerful LRM architectures.
Additionally, we evaluate our CoT-Kinetics energy modeling under multilingual settings using MGSM subsets (English, German, French, Chinese) to simulate real-world deployment scenarios, with results summarized in Table~\ref{tab:mgsm_multilingual}.
We analyze these results in detail below.
\begin{table}[t]
\centering
\normalsize
\setlength{\tabcolsep}{2pt}
\renewcommand{\arraystretch}{1.1}
\caption{GSM8K results on larger models (Higher AUROC and AUPR indicate better performance; Lower FPR@95 indicates better performance)}
\resizebox{\textwidth}{!}{%
\begin{tabular}{ll|*{9}{c}}
\toprule
\multirow{2}{*}{Dataset} & \multirow{2}{*}{Method} 
& \multicolumn{3}{c}{\textbf{R1-Qwen-32B}} 
& \multicolumn{3}{c}{\textbf{QwQ-32B}} 
& \multicolumn{3}{c}{\textbf{R1-LLaMA-70B}} \\
\cmidrule(lr){3-5} \cmidrule(lr){6-8} \cmidrule(lr){9-11}
& & AUROC~↑ & AUPR~↑ & FPR@95~↓ 
  & AUROC~↑ & AUPR~↑ & FPR@95~↓ 
  & AUROC~↑ & AUPR~↑ & FPR@95~↓ \\
\midrule

\multirow{7}{*}{GSM8K}
& Random         & 46.97 & 94.05 & 93.59 & 51.79 & 96.55 & 93.88 & 48.87 & 94.32 & 97.22 \\
& MaxProb   &54.66&94.88&94.81&66.18&97.07&72.34&60.09&96.14&85.71\\
& PPL       &55.07&94.94&93.51&66.29&97.12&72.34&60.02&96.13&85.71\\
& Entropy   &54.86&94.88&93.51&66.52&97.32&70.21&60.57&96.24&85.71\\
& CoE-R     &62.51&96.62&96.10&66.93&97.64&85.11&61.55&96.33&88.57\\
& CoE-C     &62.04&96.49&96.10&67.38&97.69&87.23&35.89&92.45&98.57\\
& CoT-Kinetics 
  & \cellcolor{blue!6}\textbf{64.33} & \cellcolor{blue!6}\textbf{96.75} & \cellcolor{blue!6}96.10
  & \cellcolor{blue!6}\textbf{70.11} & \cellcolor{blue!6}\textbf{97.79} & \cellcolor{blue!6}\textbf{70.21} 
  & \cellcolor{blue!6}\underline{61.32} & \cellcolor{blue!6}96.23 & \cellcolor{blue!6}\textbf{85.71} \\

\bottomrule
\end{tabular}
}
\label{tab:gsm8k_large_models}
\end{table}

\begin{table}[t]
\centering
\normalsize
\setlength{\tabcolsep}{2pt}
\renewcommand{\arraystretch}{1.1}
\caption{Results on MGSM dataset across four languages using DeepSeek-R1-Distill-Qwen-7B (Arrows indicate direction of improvement)}
\resizebox{\textwidth}{!}{%
\begin{tabular}{ll*{12}{c}}
\toprule
\multirow{2}{*}{Model} & \multirow{2}{*}{Method} 
& \multicolumn{3}{c}{\textbf{English (en)}} 
& \multicolumn{3}{c}{\textbf{German (de)}} 
& \multicolumn{3}{c}{\textbf{French (fr)}} 
& \multicolumn{3}{c}{\textbf{Chinese (zh)}} \\
\cmidrule(lr){3-5} \cmidrule(lr){6-8} \cmidrule(lr){9-11} \cmidrule(lr){12-14}
& & AUROC~↑ & AUPR~↑ & FPR@95~↓ 
  & AUROC~↑ & AUPR~↑ & FPR@95~↓ 
  & AUROC~↑ & AUPR~↑ & FPR@95~↓ 
  & AUROC~↑ & AUPR~↑ & FPR@95~↓ \\
\midrule

\multirow{7}{*}{\makecell[l]{DeepSeek-R1-\\Distill-Qwen-7B}} 
& Random         & 54.12 & 91.52 & 100.00   & 45.08 & 74.66 & 96.83 & 52.67 & 78.79 & 95.12 & 49.42 & 68.94 & 95.89  \\
& MaxProb &70.95&95.00&72.00&69.15&86.51&80.95&70.63&82.19&73.97& 62.76 &89.03&95.12 \\
& PPL     &70.92&94.98&72.00&69.36&86.62&80.95  &70.91&82.86&73.97&63.25&89.01&92.68  \\
& Entropy &73.24&95.40&68.00&69.15&87.04& 80.95 &70.27&82.04&72.60&65.14&90.13&92.68  \\
& CoE-R   &67.75&94.36&72.00&60.56&79.02& 80.95 &55.17&70.96&94.52&64.45&89.11&92.68 \\
& CoE-C   &66.58&94.04&72.00&58.18&78.08& 85.71 &54.15&70.21&98.63&63.43&88.61&95.12  \\

\rowcolor{blue!6}
& CoT-Kinetics & \textbf{74.03} & \textbf{95.27} & \textbf{64.00} & \textbf{69.44} & \underline{86.91} & \textbf{79.37} & \textbf{71.32} & \underline{82.36} & \textbf{72.60} & \textbf{71.04} & \textbf{91.28} & \textbf{92.68} \\

\bottomrule
\end{tabular}
}
\label{tab:mgsm_multilingual}
\end{table}

\paragraph{Generalization Ability:}
Table~\ref{tab:main_results_3bench} reports the performance of our method across vertical-domain reasoning tasks of varying difficulty. For mathematical reasoning, CoT-Kinetics consistently achieves the highest improvements (+8.5\%, +2.5\%, +10.8\%) across both the relatively easier GSM8K and the more challenging college-level TheoremQA benchmarks. For commonsense reasoning, our CoT-Kinetics modeling approach  also demonstrates the strongest scoring capability across different model scales.

For multi-task scenarios that require both world knowledge and complex problem-solving skills, we report the performance of our approach on MMLU and MMLU-Pro in Table~\ref{tab:mmlu}. These two massive multi-task benchmarks, each of which covers over 50 diverse downstream tasks, and we report the aggregated overall performance across all tasks. our CoT-Kinetics method shows strong generalization ability (+3.3\%, +5.7\%, +2.3\%) across all model scales on these challenging evaluations. Furthermore, in Table~\ref{tab:mmlu} we observe a consistent trend in the AUPR scores—emphasizing the precision of positive (correct) predictions—across different task difficulties and model sizes. As the difficulty increases from GSM8K to MMLU-Pro and model size varies, the AUPR scores of our CoT-Kinetics  method behave in a statistically consistent manner: Higher AUPR scores (from 33.39 to 48.57 on MMLU-Pro and from 76.47 to 97.84 on MMLU) are observed as the task difficulty decreases or the model scale increases. This trend suggests that our method maintains stable and reliable behavior in diverse and realistic reasoning scenarios.

Beyond the results directly demonstrating the generality, the experiment in Figure~\ref{fig:cot-kinetics-eval} (Right part) conducts a more subtle comparison by analyzing the ROC curves against selected baselines. Rather than merely observing overall performance, this deeper analysis reveals that CoT-Kinetics consistently achieves lower FPR at the same TPR, indicating that it can maintain the same level of correct reasoning identification  and even more effectively rejecting faulty outputs—a distinctive advantage not exhibited by other methods.

\textbf{Scalability:}
In Table~\ref{tab:gsm8k_large_models}, we further report the performance of our approach on three stronger heterogeneous LRMs with model sizes exceeding 30B parameters. These large-scale models are widely adopted in high-performance and reliability-critical real-world scenarios. Our CoT-Kinetics method consistently achieves the best or second-best scoring performance across all three models, demonstrating its strong capability even at larger model scales and under increasingly complex reasoning dynamics.

\textbf{Multi-lingual Applicability:}
To evaluate the cross-lingual generalization of our CoT-Kinetics energy modeling approach, we conduct experiments on multilingual settings using MGSM subsets. As shown in Table~\ref{tab:mgsm_multilingual}, CoT-Kinetics consistently achieves the best performance across four popular languages: English, German, French, and Chinese. Notably, for certain languages such as Chinese, our CoT-Kinetics even exhibits greater improvements than those observed in English. These results suggest that our solution is capable of capturing semantically coherent reasoning dynamics across different linguistic structures.

\subsection{Ablation Study}
We further investigate how alternative design choices in our CoT-Kinetics modeling affect its performance. 
We focus on two key components: (i) the aggregation strategy for reasoning-relevant tokens, and (ii) the construction of semantic action components. 
These components are critical because they directly determine how well the semantic trajectory is captured and how accurately the energy score reflects reasoning quality.

\textbf{Reasoning Token Aggregation Strategies:}
Here, we investigate three alternative strategies to the mean pooling (i.e., selecting and aggregating the hidden representations over the reasoning tokens) introduced in Section~\ref{sssec:formulCoTK} for constructing the CoT-Kinetics energy in Eq.~(\ref{eq:cot_kinetics_energy}).
The first alternative way computes the mean hidden state across all generated reasoning tokens at each layer, using this mean as the semantic center of the model's ongoing reasoning process. 
The second alternative way directly uses the hidden state of the last generated CoT token, assuming it encapsulates the accumulated reasoning information up to that point. 
The third alternative way computes the mean hidden state over all output tokens generated from both reasoning phase and the final answer conclusion phase.
Comparing these strategies aims to understand how different token aggregation schemes affect the fidelity of the modeled reasoning trajectory and ultimately the effectiveness of CoT-Kinetics energy in capturing reasoning soundness.
Our ablation results in Table \ref{tab:ablation_reasoning_token_selection} show that mean pooling over reasoning tokens is the most effective aggregation strategy, capturing the core reasoning dynamics more reliably than the alternatives do. 

\begin{table*}[t]
\centering
\normalsize
\setlength{\tabcolsep}{3pt}
\renewcommand{\arraystretch}{1.1}
\caption{
Ablation study showing that our choice best captures reasoning dynamics across 4 models.
}
\resizebox{1\textwidth}{!}{%
\begin{tabular}{l|*{12}{c}}
\toprule
\multirow{2}{*}{Strategy} 
& \multicolumn{3}{c}{\textbf{R1-Qwen-1.5B}} 
& \multicolumn{3}{c}{\textbf{R1-Qwen-7B}} 
& \multicolumn{3}{c}{\textbf{R1-LLaMA-8B}} 
& \multicolumn{3}{c}{\textbf{R1-Qwen-14B}} \\
\cmidrule(lr){2-4} \cmidrule(lr){5-7} \cmidrule(lr){8-10} \cmidrule(lr){11-13}
& AUROC~↑ & AUPR~↑ & FPR@95~↓ 
& AUROC~↑ & AUPR~↑ & FPR@95~↓ 
& AUROC~↑ & AUPR~↑ & FPR@95~↓ 
& AUROC~↑ & AUPR~↑ & FPR@95~↓ \\
\midrule

\rowcolor{blue!6}
Mean Pooling~\raisebox{-0.3\height}{\includegraphics[width=1em]{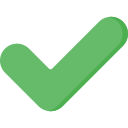}}
& \textbf{77.28} & \textbf{91.42} & \textbf{72.20} 
& \textbf{76.32} & \textbf{94.73}& \textbf{67.27} 
& \textbf{68.12} & \textbf{89.37} & \textbf{80.31} 
& \textbf{61.64} & \textbf{94.04} & \textbf{76.19} \\

Last CoT token
& 71.54 & 90.12 & 76.77 
& 71.94 & 93.61 & 73.33 
& 63.44 & 88.14 & 86.54 
& 59.27 & 92.58 & 81.28 \\

Mean of all output tokens
& 70.53 & 88.56 & 79.83 
& 70.90 & 92.96 & 75.76 
& 61.43 & 86.57 & 85.42 
& 55.64 & 91.79 & 79.58 \\

\bottomrule
\end{tabular}
}
\label{tab:ablation_reasoning_token_selection}
\end{table*}

\begin{wraptable}{r}{0.45\textwidth}
\centering
\vspace{-1.8em} 
\caption{Ablation study of CoT-Kinetics energy components}
\resizebox{\linewidth}{!}{
\begin{tabular}{ccc|ccc}
\toprule
\(\tau^i\) & \(\kappa^i\) & \(\mathcal{H}(p)\) & AUROC~↑ & AUPR~↑ & FPR@95~↓ \\
\midrule
\ding{51} &            &            & 74.35 & 92.60 & 72.12 \\
          & \ding{51}  &            & 74.07 & 92.58 & 73.33 \\
\ding{51} & \ding{51}  &            & 75.23 & 93.60 & 72.73 \\
\ding{51} &            & \ding{51}  & 69.71 & 93.00 & 78.18 \\
          & \ding{51}  & \ding{51}  & 70.11 & 93.07 & 78.18 \\
\rowcolor{blue!6}
\ding{51} & \ding{51}  & \ding{51}  & \textbf{76.32} & \textbf{94.73} & \textbf{67.27} \\
\bottomrule
\end{tabular}
}
\label{tab:ablation_semantic_energy_components}
\vspace{-1em} 
\end{wraptable}

\paragraph{CoT-Kinetics Energy Components:}
We also conduct an ablation study to investigate the contribution of each component in the semantic action formulation, including the semantic momentum energy \(\tau^i\), semantic curvature energy \(\kappa^i\), and entropy term \(\mathcal{H}(p)\), as defined in Section~\ref{sssec:formulCoTK}. Specifically, we evaluate how different combinations of these terms affect the overall scoring capability of CoT-Kinetics. Table~\ref{tab:ablation_semantic_energy_components} summarizes the ablation results on R1-7B. The results highlight that incorporating both the momentum and curvature terms significantly improves the performance over just using part of them individually. Furthermore, the results also justify the necessity of introducing the entropy term to CoT-Kinetics energy equation, which leads to the best overall assessment quality.
Moreover, all three components of the CoT-Kinetics energy (i.e., semantic momentum, semantic curvature, and entropy) are imperative for optimal performance outlined in and Table \ref{tab:ablation_semantic_energy_components}, with each contributing essential information to the evaluation of reasoning quality.

\vspace*{-1em}
\section{Conclusion}
\vspace*{-1em}
In summary, in this work, the proposed modeling approach introduces a principled theoretical framework, i.e., \emph{CoT-Kinetics} energy, to intrinsically evaluate the soundness of reasoning process within LRMs. Comparing the state-of-the-art methods, the proposed CoT-Kinetics energy score accurately reflects the casual relationship of the reasoning trajectory to the derived final answer, thus provides an effective way to better assess the quality of overall output answers in LRM. In addition to the tangible results, the proposed CoT-Kinetics energy shows the potential to assist LRMs to build a feedback loop to improve its reasoning process based on our CoT-Kinetic energy score. This is already planned as our future work undergoing.

\newpage
{
\small
\bibliographystyle{ieeetr}
\bibliography{ref}
}

\newpage
\section{Technical Appendices and Supplementary Material}
\subsection{Limitation}

Although we have evaluated CoT-Kinetics on a wide range of state-of-the-art LRMs, the availability of open-sourced models remains limited, particularly for extremely large-scale LRMs. We hope that more high-quality open-sourced LRMs will become available in the future to facilitate broader exploration and validation of CoT-Kinetics at even larger model scales.
\subsection{Related Work}
\label{rel}
\textbf{Large Reasoning Models (LRMs):} 
With the advancement of deep learning \cite{lu2019singleimagesuperresolution,Lu_2022,Lu_2023,chen2025fedbipheterogeneousoneshotfederated,li2024towards,li2024distinct,liu2024multimodalpragmaticjailbreaktexttoimage,rong2025can,zhong2025enhancing,cui2024correlation,jing2023category,jing2023multimodal,lu2023tf,lu2024mace,lu2024robust,ye2023advancingfederatedlearning6g,DBLP:conf/globecom/YeAYC23, Wan_EpiODE_ICML25,Wan_GraphODE_ICML25,FedTGE_ICLR25,S3GCL_ICML24,FGGP_AAAI24,FedSSP_NeurIPS,FGSSL_IJCAI23,Epi_Survey_KDD24,zhou2025opendrivevlaendtoendautonomousdriving,zhang2024webpilotversatileautonomousmultiagent,liu2024perft}, Large Language Models (LLMs) \cite{huang2025keeping,Huang_2025,Lorasculpt_CVPR25,NRCA_ICML25,yang2025lighthouselanguageenhancingllm}have achieved significant progress across multiple domains \cite{10966003,zhou2025learning,chen2023feddatapproachfoundationmodel,chen2023theoremqatheoremdrivenquestionanswering,bi2024visual,wang2024conuconformaluncertaintylarge,wang2024wordsequenceentropyuncertaintyestimation,wang2025sconuselectiveconformaluncertainty,liu2025focalpoenhancingpreferenceoptimizing,wang2025fine,zhang2023spotrevisitingvideolanguagemodels,ye2025survey}.
Comparing with LLMs, LRMs have demonstrated that extending the reasoning process can substantially enhance model inference performance \cite{cui2025drawthoughtunleashingmultimodal,yuan2025reversalthoughtenhancinglarge}. 
For example, \texttt{OpenAI-o1}~\cite{openai2024openaio1card}, \texttt{Qwen-QwQ}~\cite{qwq32b} and \texttt{DeepSeek-R1}~\cite{deepseekai2025deepseekr1incentivizingreasoningcapability} explicitly perform CoT reasoning by constructing an intermediate reasoning trajectory trying to justify the final answer~\cite{wei2023chainofthoughtpromptingelicitsreasoning}. LRMs show to be effective in domains requiring complex logical deduction, such as mathematics and programming~\cite{chen2023theoremqatheoremdrivenquestionanswering,cobbe2021gsm8k,hendrycks2021measuringmassivemultitasklanguage,shi2022languagemodelsmultilingualchainofthought,wang2024mmluprorobustchallengingmultitask,talmor-etal-2019-commonsenseqa,wang2025visuothinkempoweringlvlmreasoning,zhang2023babyscothoughtleveraginglarge,yue2025thinkhierarchicallyactdynamically,zhao2025chartcoder}.
Unlike conventional LLMs, which produce an answer in a single forward pass guided by shallow heuristics, LRMs follow an explicit “think‑then‑answer” routine~\cite{deepseekai2025deepseekr1incentivizingreasoningcapability,openai2024openaio1card,qwq32b,geminiteam2025geminifamilyhighlycapable,muennighoff2025s1simpletesttimescaling}. They first enclose a CoT segment between special tokens (e.g., \texttt{<think>}…\texttt{</think>}) and repeatedly update internal representations while attending to these intermediate tokens; They generate the final answer only after completing the step-by-step deliberation process.

Current efforts to evaluate the overall output quality of an LRM can be broadly categorized into external and internal assessment methods, which are separately reviewed as follows.

\textbf{External Assessment:}
This type of assessment methods relies on supervision signals external to the model, such as linear probing techniques~\citep{wang2025thoughtprobe}, additional annotations or knowledge bases~\citep{li2023makinglargelanguagemodels, nguyen2024directevaluationchainofthoughtmultihop}, and proxy models~\citep{he2025largelanguagemodelsdetect}.
For example, \cite{kadavath2022languagemodelsmostlyknow} leverages the instruction-following capabilities of LLMs, enhanced by RLHF~\citep{ouyang2022traininglanguagemodelsfollow}, to generate confidence scores through carefully crafted prompts.
This paradigm encompasses a variety of general methodologies that directly elicit confidence estimates from LLMs via multi-stage pipelines, as broadly evaluated in~\citep{lin2022teachingmodelsexpressuncertainty, tian2023justaskcalibrationstrategies}.
\cite{manakul2023selfcheckgptzeroresourceblackboxhallucination} further investigates this approach by prompting GPT-3 to generate biographical passages from the WikiBio dataset and manually annotating the factuality of the generated content.
Similarly, the PSA pipeline~\citep{xiong2024llmsexpressuncertaintyempirical} estimates confidence by perturbing prompts~\citep{gao2024spuqperturbationbaseduncertaintyquantification} to produce multiple outputs and subsequently measuring their consistency.
However, these methods typically require access to ground-truth labels, external resources, or auxiliary models, limiting their generalizability in real-world scenarios where such information may be unavailable or impractical to obtain. This limitation motivates the exploration of a label-free, model-internal evaluation strategies.

\textbf{Internal Assessment:}
This type of assessment methods leverages the internal states obtained from the model itself, such as hidden representations~\citep{wang2025latentspacechainofembeddingenables} or uncertainty-based signals~\citep{shih2023longhorizontemperaturescaling, Huang_2025, si2023promptinggpt3reliable}, to infer the reliability of its outputs without relying on external supervision.
For example, \cite{plaut2025probabilitieschatllmsmiscalibrated} uses softmax probabilities to predict the correctness of multiple-choice questions, while \cite{hendrycks2022scalingoutofdistributiondetectionrealworld} explores temperature scaling techniques for output probabilities.
\cite{kuhn2023semanticuncertaintylinguisticinvariances} and \cite{farquhar2024detecting} investigate the concept of semantic entropy, aiming to capture uncertainty beyond surface-level probabilities.
Furthermore, \cite{duan2024shiftingattentionrelevancepredictive} introduces a shift toward utilizing semantic information in entropy-based reliability estimation.
Notably, these approaches primarily focus on detecting specific types of factual errors, particularly hallucinations.

While this direction alleviates the dependence on external labels and models, most existing internal assessment approaches still evaluate the output holistically, without explicitly disentangling the reasoning trajectory from the final answer. As a result, the correctness of the final answer inevitably introduces bias, masking deficiencies in the reasoning process itself. Consequently, these methods fail to properly assess the intrinsic quality of the model's reasoning, leaving a critical gap that motivates the need for more fine-grained evaluation frameworks.

\textbf{Summary:}
Comparing to the two types of existing methods, our {CoT-Kinetics} energy modeling approach improves the state-of-the-art solutions on the following two key aspects. 
First of all, CoT-Kinetic energy extracts the reasoning part out of the overall output answer and can explicitly assess the soundness of the reasoning part without any external information. Note that previous studies~\citep{wang2024chain, turpin2023languagemodelsdontsay,lyu2023faithfulchainofthoughtreasoning} have shown that flawed reasoning can yield correct answers. Hence, the capability able to independently assess the reasoning part is crucial~\citep{nguyen2024directevaluationchainofthoughtmultihop}, while none of the existing methods can do that. 
Second, techniques used to establish the CoT-Kinetics energy indeed overcome the deficiency of the existing modeling approaches. CoT-Kinetics energy offers a more principled modeling of the reasoning process and  gives a logical scoring capability to assess the overall quality of the overall output answers, surpassing all existing methods substantially.




\subsection{Benchmark Dataset Details}

\label{appendix:dataset_details}

We evaluate the generality and effectiveness of our proposed CoT-Kinetics method across six widely adopted reasoning benchmarks. Each dataset is brief

\textbf{GSM8K~\cite{cobbe2021gsm8k}:}
a dataset designed for math word problem-solving tasks, involving arithmetic calculations and logical reasoning. Each question expects an integer as the final answer, accompanied by explicit reasoning steps. We follow standard evaluation metrics and report accuracy.

\textbf{CommonsenseQA~\cite{talmor-etal-2019-commonsenseqa}:} a multiple-choice dataset evaluating commonsense reasoning. Each question provides five candidate answers (A–E), requiring the model to choose the correct option. Answers are evaluated by accuracy, emphasizing both factual knowledge and reasoning capability.

\textbf{TheoremQA~\cite{chen2023theoremqatheoremdrivenquestionanswering}:} a dataset used to evaluate mathematical theorem reasoning and deduction abilities. It includes questions covering diverse mathematical topics, with answers in various formats such as numerical results, Boolean values, or lists of numbers. Performance is measured via accuracy.

\textbf{MMLU~\cite{hendrycks2021measuringmassivemultitasklanguage}:} a dataset assessing models across diverse knowledge domains. It comprises multiple-choice questions from fields such as mathematics, sciences, and humanities, making it ideal for evaluating model versatility and general knowledge retrieval. Answers are evaluated using accuracy.

\textbf{MMLU-Pro~\cite{wang2024mmluprorobustchallengingmultitask}:} An extension of MMLU, MMLU-Pro introduces a more challenging set of questions to test robust multitask reasoning capabilities of LLMs. Similar to MMLU, questions span multiple disciplines and use accuracy as the evaluation metric.

\textbf{MGSM~\cite{shi2022languagemodelsmultilingualchainofthought}}
MGSM is a multilingual benchmark focusing on math word problems. It evaluates arithmetic reasoning capabilities across multiple languages, including English, German, French, and Chinese. We evaluate performance using accuracy on numeric answers, demonstrating multilingual applicability.

Detailed dataset statistics, task specifications, prompting templates, and illustrative examples for these benchmarks are provided in Table~\ref{tab:downstream_datasets} and Table~\ref{tab:evaluation_datasets}.

\begin{table}[t]
    \centering
    \caption{Details of the downstream reasoning datasets (vertical-domain)}
    \renewcommand{\arraystretch}{1.3} 
    \setlength{\tabcolsep}{4pt} 
    \resizebox{1\textwidth}{!}{
    \begin{tabular}{
        >{\centering\arraybackslash}m{2.8cm}  
        >{\centering\arraybackslash}p{5.2cm} 
        >{\centering\arraybackslash}p{5.2cm} 
        >{\centering\arraybackslash}p{5.2cm}
    }
    \toprule
    \makecell{\textbf{Datasets} \\ (Train/Test)} & \makecell{\textbf{GSM8K} \\ (7473/1319)} & \makecell{\textbf{CommonsenseQA} \\ (9741/1140)} & \makecell{\textbf{TheoremQA} \\ (-/800)} \\
    \midrule
    \textbf{Venue} & \cite{cobbe2021gsm8k} & \cite{talmor-etal-2019-commonsenseqa} & \cite{chen2023theoremqatheoremdrivenquestionanswering} \\
    \midrule
    \textbf{Task} & Math word problem solving & Commonsense reasoning & Mathematical theorem reasoning \\
    \midrule
    \textbf{Metric} & Accuracy (f) & Accuracy (f) & Accuracy (f) \\
    \midrule
    \textbf{Answer Format} & Number & Option & Option / Boolean / Number / List of Numbers \\
    \midrule
    \textbf{Prompt Template} &
    \begin{minipage}[t]{\linewidth}
    \footnotesize
    Solve this math problem. Provide reasoning steps before giving the final answer in the format ``Answer:''. Only output the integer after ``Answer:''.\\[0.5em]
    \textbf{Question:} \{input\_data\}
    \end{minipage}
    &
    \begin{minipage}[t]{\linewidth}
    \footnotesize
    Answer the multiple choice commonsense question. Final line format: ``Answer: \$LETTER'' (A/B/C/D/E). Provide step-by-step reasoning.\\[0.5em]
    \textbf{Question:} \{input\_data\}
    \end{minipage}
    &
    \begin{minipage}[t]{\linewidth}
    \footnotesize
    Complete the math problem considering the provided answer type. Different formats are required for boolean, integer, list of numbers, or option.\\[0.5em]
    \textbf{Question:} \{input\_data\}\\
    \textbf{Answer Type:} \{answer\_type\}
    \end{minipage}
    \\
    \midrule
    \textbf{Example} &
    \begin{minipage}[t]{\linewidth}
    \footnotesize
    \textbf{Q}: Natalia sold clips to 48 friends in April, and half as many in May. How many in total?\\
    \textbf{A}: 72
    \end{minipage}
    &
    \begin{minipage}[t]{\linewidth}
    \footnotesize
    \textbf{Q}: The sanctions against the school were a punishing blow. They seemed to what the efforts the school made to change?\\
    Options: A. ignore B. enforce C. authoritarian D. yell at E. avoid\\
    \textbf{A}: A
    \end{minipage}
    &
    \begin{minipage}[t]{\linewidth}
    \footnotesize
    \textbf{Q}: How many ways are there to divide a set of 8 elements into 5 non-empty ordered subsets?\\
    \textbf{A}: 11760
    \end{minipage}
    \\
    \bottomrule
    \end{tabular}
    }
    \label{tab:downstream_datasets}
\end{table}

\begin{table}[t]
    \centering
    \caption{Details of the multi-domain and multilingual evaluation datasets
}
    \renewcommand{\arraystretch}{1.2} 
    \setlength{\tabcolsep}{4pt} 
    \resizebox{1\textwidth}{!}{
    \begin{tabular}{
        >{\centering\arraybackslash}m{2.8cm} 
        >{\centering\arraybackslash}p{5.2cm} 
        >{\centering\arraybackslash}p{5.2cm} 
        >{\centering\arraybackslash}p{5.2cm}
    }
    \toprule
    \makecell{\textbf{Datasets} \\ (Train/Test)} & \makecell{\textbf{MMLU} \\ (99842/14042)} & \makecell{\textbf{MMLU-Pro} \\ (-/12032)} & \makecell{\textbf{MGSM} \\ (-/250 per language)} \\
    \midrule
    \textbf{Venue} & \cite{hendrycks2021measuringmassivemultitasklanguage} & \cite{wang2024mmluprorobustchallengingmultitask} & \cite{shi2022languagemodelsmultilingualchainofthought} \\
    \midrule
    \textbf{Task} & Multitask multiple-choice QA & Robust multitask multiple-choice QA & Multilingual math word problem solving \\
    \midrule
    \textbf{Metric} & Accuracy (f) & Accuracy (f) & Accuracy (f) \\
    \midrule
    \textbf{Answer Format} & Option & Option & Number \\
    \midrule
    \textbf{Prompt Template} &
    \begin{minipage}[t]{\linewidth}
    \footnotesize
    Answer the multiple choice question. Final line format: ``Answer: \$LETTER''.\\[0.5em]
    \textbf{Question:} \{input\_data\}
    \end{minipage}
    &
    \begin{minipage}[t]{\linewidth}
    \footnotesize
    Same as MMLU.\\[0.5em]
    \textbf{Question:} \{input\_data\}
    \end{minipage}
    &
    \begin{minipage}[t]{\linewidth}
    \footnotesize
    Solve the math problem. Output final answer after ``Answer:''.\\[0.5em]
    \textbf{Question:} \{input\_data\}
    \end{minipage}
    \\
    \midrule
    \textbf{Example} &
    \begin{minipage}[t]{\linewidth}
    \footnotesize
    \textbf{Q}: Let \( p = (1,2,5,4)(2,3) \) in \( S_5 \). Find the index of \( \langle p \rangle \) in \( S_5 \).\\
    Options: A. 8 \quad B. 2 \quad C. 24 \quad D. 120\\
    \textbf{A}: C
    \end{minipage}
    &
    \begin{minipage}[t]{\linewidth}
    \footnotesize
    \textbf{Q}: Find the characteristic of the ring \(2\mathbb{Z}\).\\
    Options: A. 0 \quad B. 30 \quad C. 3 \quad D. 10 \quad E. 12 \quad F. 50 \quad G. 2 \quad H. 100 \quad I. 20 \quad J. 5\\
    \textbf{A}: A
    \end{minipage}
    &
    \begin{minipage}[t]{\linewidth}
    \footnotesize
    \textbf{Q}: A robe takes 2 bolts of blue fiber and half that much white fiber. How many bolts in total?\\
    \textbf{A}: 3
    \end{minipage}
    \\
    \bottomrule
    \end{tabular}
    }
    \label{tab:evaluation_datasets}
\end{table}

\subsection{Model Details}
\label{appendix:model-configurations}

Table~\ref{tab:model_configurations} summarizes the model configurations used in our experiments.
We cover a range of model scales and architectures to validate the robustness and generality of CoT-Kinetics across diverse LLM backbones.
The models differ in their base architectures, number of layers, hidden dimensions, and reserved special tokens (\texttt{<think>}, \texttt{</think>}) for explicit reasoning phase delineation during inference.

\begin{table}[t]
    \centering
    \caption{Model configuration details used in our experiments.}
    \renewcommand{\arraystretch}{1.3} 
    \setlength{\tabcolsep}{-7pt} 
    \resizebox{0.97\textwidth}{!}{
    \begin{tabular}{
        >{\centering\arraybackslash}m{5cm}
        >{\centering\arraybackslash}m{4.2cm}
        >{\centering\arraybackslash}m{4.2cm}
        >{\centering\arraybackslash}m{4.2cm}
        >{\raggedright\arraybackslash}m{4.2cm}
    }
    \toprule
    \textbf{Model Name} & \textbf{Base Model} & \textbf{Number of Layers} & \textbf{Hidden Size} & \textbf{Special Token IDs} \\
    \midrule
    DeepSeek-R1-Distill-Qwen-1.5B & Qwen2.5-Math-1.5B & 28 & 1536 & \texttt{<think>}: 151648\newline \texttt{</think>}: 151649 \\
    \midrule
    DeepSeek-R1-Distill-Qwen-7B & Qwen2.5-Math-7B & 28 & 3584 & \texttt{<think>}: 151648\newline \texttt{</think>}: 151649 \\
    \midrule
    DeepSeek-R1-Distill-Llama-8B & Llama-3.1-8B & 32 & 4096 & \texttt{<think>}: 128013\newline \texttt{</think>}: 128014 \\
    \midrule
    DeepSeek-R1-Distill-Qwen-14B & Qwen2.5-14B & 48 & 5120 & \texttt{<think>}: 151648\newline \texttt{</think>}: 151649 \\
    \midrule
    DeepSeek-R1-Distill-Qwen-32B & Qwen2.5-32B & 64 & 5120 & \texttt{<think>}: 151648\newline \texttt{</think>}: 151649 \\
    \midrule
    DeepSeek-R1-Distill-Llama-70B & Llama-3.3-70B-Instruct & 80 & 8192 & \texttt{<think>}: 128013\newline \texttt{</think>}: 128014 \\
    \midrule
    QwQ-32B & Qwen2.5-32B & 64 & 5120 & \texttt{<think>}: 151667\newline \texttt{</think>}: 151668 \\
    \bottomrule
    \end{tabular}
    }
    \label{tab:model_configurations}
\end{table}

\subsection{Detailed Mathematical Derivation of Evaluation Metrics}
\label{appendix:evaluation_metrics}

In this appendix, we provide the formal definitions and detailed derivations of the evaluation metrics used in this paper, namely AUROC, AUPR, and FPR@95, for assessing the alignment between CoT-Kinetics energy scores \( \mathcal{E}_\text{CoT} \) and the correctness of answers.

\subsubsection{Problem Setup and Notations}

We assume a dataset consisting of \(B\) samples \(\{(Q_i, \mathcal{E}_i, A_i^{\text{GT}})\}_{i=1}^{B}\), where:
\begin{itemize}
    \item \(Q_i\) denotes the \(i\)-th query.
    \item \(\mathcal{E}_i\) denotes the CoT-Kinetics energy score computed by our method for sample \(i\).
    \item \(A_i^{\text{GT}} \in \{0,1\}\) is the ground-truth correctness label (\(1\) for correct, \(0\) for incorrect).
\end{itemize}

We further define:
\begin{itemize}
    \item \(\mathcal{P} = \{i \mid A_i^{\text{GT}} = 1\}\) : set of indices for positive (correct) samples.
    \item \(\mathcal{N} = \{j \mid A_j^{\text{GT}} = 0\}\) : set of indices for negative (incorrect) samples.
\end{itemize}

\subsubsection{AUROC (Area Under Receiver Operating Characteristic)}

AUROC measures the probability that a randomly chosen positive sample is ranked higher than a randomly chosen negative sample based on the CoT-Kinetics energy scores. It is formally computed as:

\begin{equation}
    \text{AUROC} = \frac{1}{|\mathcal{P}||\mathcal{N}|} 
    \sum_{i \in \mathcal{P}} \sum_{j \in \mathcal{N}} 
    \mathbb{1}(\mathcal{E}_i > \mathcal{E}_j),
\end{equation}
where \(\mathbb{1}(\cdot)\) is the indicator function:
\begin{equation}
\mathbb{1}(\mathcal{E}_i > \mathcal{E}_j) = 
\begin{cases}
1, & \text{if } \mathcal{E}_i > \mathcal{E}_j, \\
0, & \text{otherwise}.
\end{cases}
\end{equation}

A higher AUROC (closer to 1.0) indicates better global ranking capability.

\subsubsection{AUPR (Area Under Precision-Recall Curve)}

AUPR quantifies the precision-recall trade-off, especially important under class imbalance. At a threshold \(t\), precision and recall are defined as:
\begin{align}
    \text{Precision}(t) &= \frac{TP(t)}{TP(t) + FP(t)}, \\[4pt]
    \text{Recall}(t) &= \frac{TP(t)}{TP(t) + FN(t)},
\end{align}
where \(TP(t)\), \(FP(t)\), and \(FN(t)\) denote true positives, false positives, and false negatives at threshold \(t\), respectively.

AUPR is computed as the area under the precision-recall curve:
\begin{equation}
    \text{AUPR} = \int_{0}^{1} \text{Precision}(\text{Recall}) \, d(\text{Recall}),
\end{equation}
which is typically approximated using numerical integration methods (e.g., trapezoidal rule).

\subsubsection{FPR@95 (False Positive Rate at 95\% Recall)}

FPR@95 measures the false positive rate at the threshold where the recall reaches exactly \(95\%\), evaluating robustness under stringent correctness requirements.

We first find the decision threshold \(t^*\) such that:
\begin{equation}
    t^* = \arg\min_{t} \left| \text{Recall}(t) - 0.95 \right|.
\end{equation}

Then, the FPR@95 is computed as:
\begin{equation}
    \text{FPR@95} = \frac{FP(t^*)}{|\mathcal{N}|}.
\end{equation}

Lower FPR@95 indicates that the CoT-Kinetics energy score reliably identifies correct reasoning trajectories with minimal false positives even at high-recall thresholds.

\end{document}